\documentclass[letterpaper, 10 pt, conference]{ieeeconf}  

\IEEEoverridecommandlockouts                              

\usepackage{graphics} 
\usepackage{epsfig} 
\usepackage{times} 
\usepackage{amsmath} 
\usepackage{amssymb}  

\usepackage{booktabs}
\usepackage{multirow}
\usepackage{graphicx}
\usepackage{paralist}
\usepackage[dvipsnames]{xcolor}
\usepackage{pifont}
\usepackage[capitalise]{cleveref}
\usepackage{caption}
\usepackage{wrapfig}
\usepackage{titletoc}
\usepackage{colortbl}

\DeclareMathOperator*{\argmax}{arg\,max}

\newcommand{\cmark}{\textcolor[HTML]{59a14f}{\ding{51}}}%
\newcommand{\xmark}{\textcolor[HTML]{e15759}{\ding{55}}}%
\newcommand{\mbo}{\mathbf{o}}%
\newcommand{\mba}{\mathbf{a}}%
\newcommand{\mbv}{\mathbf{v}}
\newcommand{\linweight}{\;{\color{red} \rule{2pt}{1pt}\,\rule{2pt}{2pt}\,\rule{2pt}{3pt}\,\rule{2pt}{4pt}}\;}
\newcommand{\lastweight}{\;{\color{blue} \rule{2pt}{1pt}\,\rule{2pt}{1pt}\,\rule{2pt}{1pt}\,\rule{2pt}{4pt}}\;}
\allowdisplaybreaks[4]

\definecolor{Gray}{gray}{0.95}
\definecolor{LightCyan}{rgb}{0.88,1,1}
\newcolumntype{a}{>{\columncolor{Gray}}r}
\newcolumntype{b}{>{\columncolor{white}}r}

\usepackage{todonotes}
\title{\LARGE \bf
Multi-Task Learning with Sequence-Conditioned Transporter Networks
}

\author{Michael H. Lim$^{1,2,*}$, Andy Zeng$^{3}$, Brian Ichter$^{3}$, Maryam Bandari$^{1}$, Erwin Coumans$^{3}$,\\
Claire Tomlin$^{2}$, Stefan Schaal$^{1}$, Aleksandra Faust$^{3}$
\thanks{$^{1}$Intrinsic, USA.}%
\thanks{$^{2}$Hybrid Systems Lab at the Unversity of California, Berkeley, USA.}%
\thanks{$^{3}$Google Research, USA.}%
\thanks{$^{*}$Work done while the author was an intern at Intrinsic.}
\thanks{Correspondence to {\tt michaelhlim@berkeley.edu}.}
}

\begin{document}

\maketitle
\thispagestyle{empty}
\pagestyle{empty}

\begin{abstract}
Enabling robots to solve multiple manipulation tasks has a wide range of industrial applications.
While learning-based approaches enjoy flexibility and generalizability, scaling these approaches to solve such compositional tasks remains a challenge.
In this work, we aim to solve multi-task learning through the lens of sequence-conditioning and weighted sampling.
First, we propose a new suite of benchmark specifically aimed at compositional tasks, MultiRavens, which allows defining custom task combinations through task modules that are inspired by industrial tasks and exemplify the difficulties in vision-based learning and planning methods.
Second, we propose a vision-based end-to-end system architecture, Sequence-Conditioned Transporter Networks, which augments Goal-Conditioned Transporter Networks with sequence-conditioning and weighted sampling and can efficiently learn to solve multi-task long horizon problems.
Our analysis suggests that not only the new framework significantly improves pick-and-place performance on novel 10 multi-task benchmark problems, but also the multi-task learning with weighted sampling can vastly improve learning and agent performances on individual tasks.
\end{abstract}

\section{Introduction}
\label{sec:intro}
Compositional multi-task settings are found in a number of industrial settings: from placing gears and chaining parts for assembly in a manufacturing plant, to routing cables and stacking servers for datacenter logistics.
Such tasks often require robots to perform multiple maneuvers using different manipulation skills \textit{(primitives)}.
Humans have the ability to generalize between compositional tasks of varying complexity and diversity \cite{compositionalRLsurvey}, but solving compositional tasks remains a major challenge in robotics.
Enabling robotic autonomous agents to learn and reason through compositional multi-tasks can further unlock autonomous manufacturing frontiers by diversifying the problems that robots can solve while significantly decreasing production costs \cite{luo2021robust}.

Complex manipulation is compositional by nature -- requiring multi-step interactions with the physical world, and can be formulated as sequencing multiple individual tasks over a longer horizon. Learning-based approaches offer the flexibility to learn complex manipulation skills and generalize to unseen configurations \cite{levine2016end,kalashnikov18aqtopt,mahler2019,florence2018dense}.
However, scaling these approaches to solve compositional tasks may not always be successful \cite{kalashnikov18aqtopt,toyer2020magical,mtopt2021arxiv}.
Specifically, task compositionality often requires exponentially increasing amounts of demonstrations for agents to learn and generalize well, while avoiding compounding long horizon planning errors.
In recent works in compositional task learning, it has been shown that learning agents should leverage shared task structures \cite{gur2021adversarial,Sohn2020Meta}, and techniques such as sequence-conditioning \cite{dasari2020transformer}, balanced sampling and learning curricula \cite{mtopt2021arxiv} can aid this process.
Thus, learning systems should effectively reason with task structures to solve compositional tasks.

In this work, we propose a new suite of benchmark problems inspired by industrial compositional tasks, \textit{MultiRavens}, which allows customizing compositional task problems with task modules.
To effectively solve these compositional tasks, we then propose a system architecture that scales up to solve long horizon compositional manipulation tasks.
Our vision-based end-to-end system architecture, \textit{Sequence-Conditioned Transporter Networks}, uses sequence-conditioning and multi-task learning with weighted sampling to solve multi-task long horizon problems, leveraging the task structure and compositionality with little data.

Our results show that the new framework significantly outperforms original Transporter networks \cite{zeng2020transporter,seita_bags_2021} on 10 multi-task benchmark problems in \textit{MultiRavens}, where each problem exemplifies the difficulties in vision-based learning and planning tasks.
Furthermore, our results suggest that multi-task learning with weighted sampling can improve learning and agent performances on individual tasks.
Beyond compositional tasks, our results provide insights for effective intermediate representations and learning shared structures.

\begin{figure}[t]
\centering
  \includegraphics[width=0.70\textwidth]{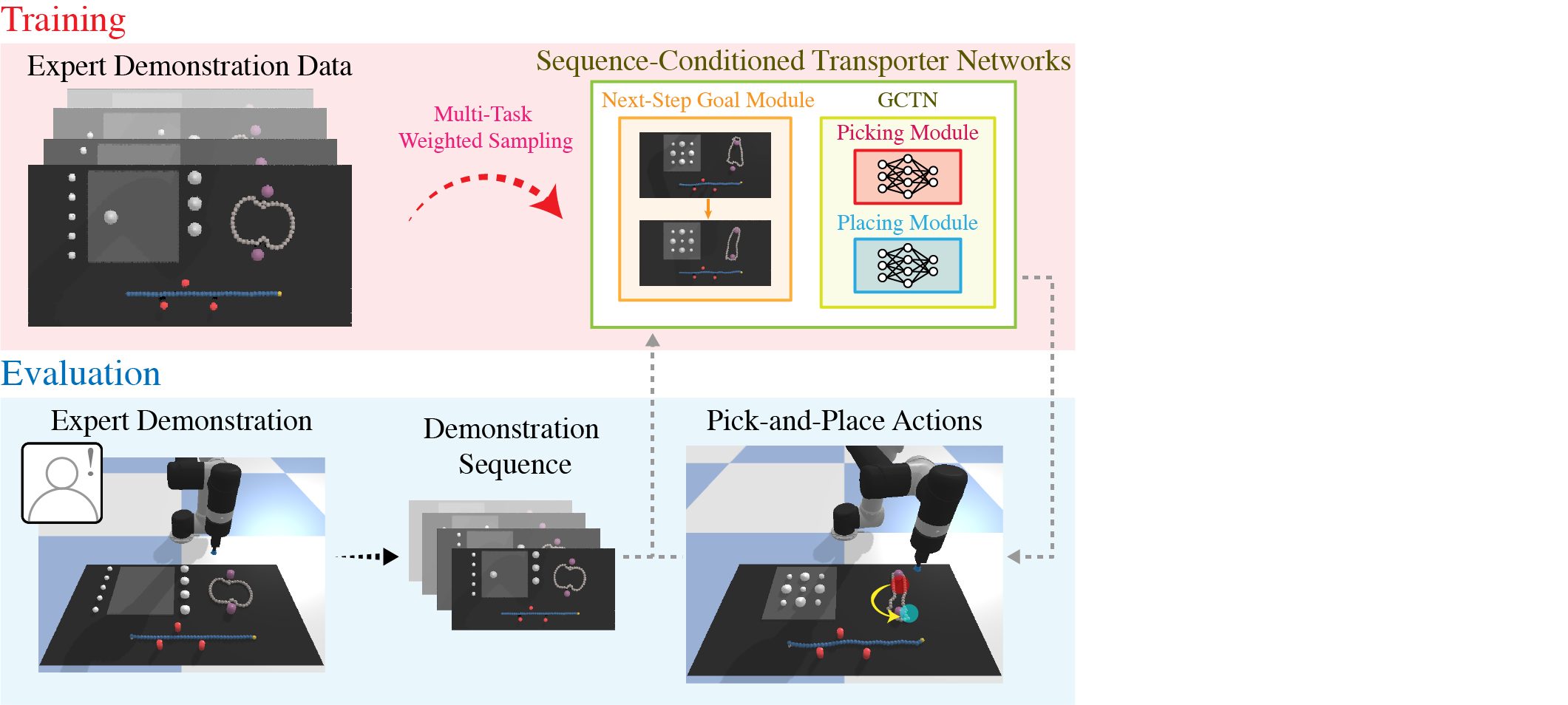}
  \caption{SCTN is trained through multi-task weighted sampling, and can reason with demonstration sequence images with the trained networks to output pick-and-place actions.}
  \vspace{-1.5em}
  \label{fig:sctn-demo}
\end{figure}

\begin{figure*}[t]
\centering
\vspace{0.0625in}
  \includegraphics[width=\textwidth]{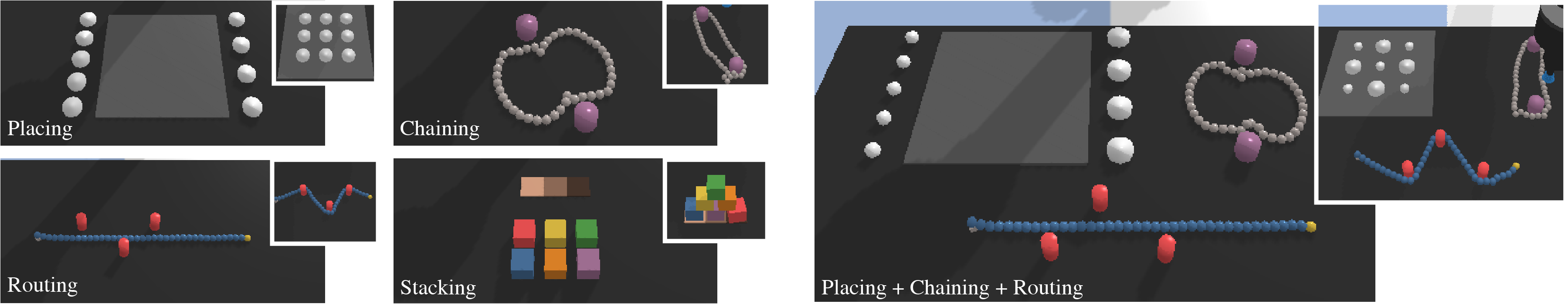}
  
  \vspace{-0.0325in}
  \caption{\textbf{MultiRavens.} Task modules introduced in MultiRavens: \textit{placing, chaining, routing} and \textit{stacking}, and their  goal configurations (left). 
  An example multi-task problem \textit{placing + chaining + routing} and its goal configuration (right).}
  \label{fig:tasks}
  \vspace{-1.5em}
  \vspace{-0.03in}
\end{figure*}
\section{Related Work}
\label{sec:related}
\subsection{Pick-and-Place Tasks}
Recent works in pick-and-place tasks leverage learned models to generate pick-and-place policies that can successfully generalize to unseen objects \cite{zeng2020transporter,seita_bags_2021,gualtieri2018pick,zakka2019form2fit,hundtgood,berscheid2020self,wu2019learning,devin2020self,khansari2020action, song2020grasping}.
These vision-based agents produce per-pixel scores in an image, which are then mapped to physical pick-and-place coordinates.
This approach is shown to be successful for tasks including robotic warehouse order picking \cite{zeng2018robotic}, pushing and grasping \cite{zeng2018learning}, tossing objects \cite{zeng2019tossingbot}, kit assembly \cite{zakka2019form2fit}, and mobile manipulation \cite{wu2020spatial}.

\subsection{Multi-Task Learning}
Multi-task learning approaches aim to improve learning efficiency through shared structure and data between tasks.
While multi-task learning and systems have shown effectiveness across several domains \cite{hundtgood}, they often require careful choices of tasks, balanced sampling, and learning curricula \cite{gur2021adversarial,Sohn2020Meta,dasari2020transformer}.
Consequently, multi-task learning may not always improve learning, sometimes diminishing performances as shown in vision-based continuous control manipulation tasks \cite{toyer2020magical}.
Moreover, composing tasks together amplifies the problem of compounding errors \cite{zhang2016queryefficient,laskey2016support,Ng00algorithmsfor} and the need for large number of demonstrations \cite{finn2017oneshot,ross2011reduction}.
Understanding task structure may be a key ingredient in scaling reinforcement learning methods to compositional tasks \cite{nachum2019nearoptimal,li2021solving,sohn2018hierarchical,sohn2019meta}, as leveraging task structures can successfully solve realistic compositional tasks including web navigation \cite{gur2021adversarial} and subtask dependency inference \cite{Sohn2020Meta}.

Sequence-conditioning combined with multi-task learning can also be viewed as a variant of one-shot meta-learning.
Meta-learning approaches aim to learn policies that can quickly adapt and generalize to new task instances \cite{finn2017oneshot}, and some of these approaches also condition the policies on demonstration videos or sequences \cite{dasari2020transformer,yu2018oneshot}.
While most prior works in meta-learning focus on generalizing to new unseen tasks of short planning horizon, our work focuses on multiple seen tasks that result in long planning horizon.

\section{Background}

\subsection{Problem Formulation}
Consider the problem of rearranging objects as learning a policy $\pi:\mathcal{O}\to\mathcal{A}$ that maps the robot visual observations $\mbo_t \in \mathcal{O}$ to corresponding pick-and-place actions $\mba_t \in \mathcal{A}$:
\begin{equation}
    \mbo_t \mapsto \pi(\mbo_t) = \mba_t = (\mathcal{T}_\text{pick}, \mathcal{T}_\text{place}) \in \mathcal{A}.
\end{equation}
Both $\mathcal{T}_\text{pick}, \mathcal{T}_\text{place} \in $ SE(2) describe the geometric pose of the end effector when grasping and releasing the object, where SE(2) is the Special Euclidean group that covers all translations and rotations in the 2D $xy$-plane.
For a given pick-and-place action pair, the end effector approaches $\mathcal{T}_\text{pick}$ until contact is detected, attempts to grasp the object, lifts up to a fixed $z$-height, approaches $\mathcal{T}_\text{place}$ to lower the object until contact is detected, and releases the grasp.
This framework can be generalized further to many tasks solvable with two-pose motion primitives \cite{zeng2020transporter}.

It is crucial that the policy $\pi$ learns from a training procedure with closed-loop visual feedback, as many compositional tasks require sequential planning over long horizons while avoiding compounding error.
The visual observations of the expert demonstrations are given from the top-down view of the planar surface, from which the picking and placing poses are sampled.
We assume access to a small dataset $\mathcal{D} = \{\zeta_1, \zeta_2, \cdots, \zeta_N\}$ of $N$ stochastic expert demonstrations, used to train $\pi$.
Each episode $\zeta_i=\{(\mbo_1, \mba_1), \cdots, (\mbo_{T_i}, \mba_{T_i})\}$ of length $T_i$ is a sequence of observation-action pairs $(\mbo_t, \mba_t)$. 

\begin{figure*}[t]
\centering
\vspace{0.0625in}
  \includegraphics[width=\textwidth]{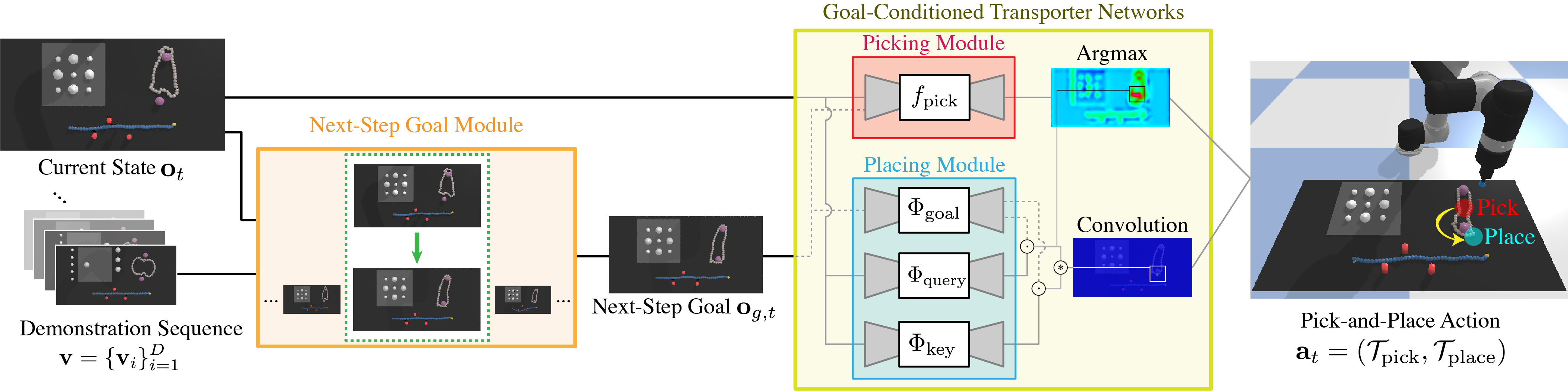}
  \caption{\textbf{Sequence-Conditioned Transporter Networks (SCTN).} Roll out of SCTN policy on \textit{placing + chaining + routing}. SCTN takes in as input the current state $\mbo_t$ and the demonstration sequence $\mbv$. The next-step goal module determines the next-step goal $\mbo_{g,t}$ from $\mbo_t$ and $\mbv$, which is then fed into the GCTN module with $\mbo_t$ to obtain the pick-and-place action $\mba_t$.}
  \vspace{-1.9em}
  \label{fig:sctn}
\end{figure*}
\subsection{Transporter Frameworks}
\textit{Transporter Networks} \cite{zeng2020transporter} and \textit{Goal-Conditioned Transporter Networks} (GCTN) \cite{seita_bags_2021} are model architectures that learn pick-and-place actions by using attention and convolution modules. 
For picking, the attention module predicts regions of placing interest given the current observation.
Then for placing, the convolution module uses the picking region to determine the spatial displacement via cross-correlating the image features over the observation and goal (for GCTN).

Both Transporter frameworks utilize the spatial equivariance of 2D visual observations, where the top-down observations are invariant under 2D translations and rotations \cite{kondor2018generalization,cohen2016group,zeng2019learning}. 
Thus, the observations $\mbo_t$ are top-down orthographic images of the workspace, where each pixel represents a narrow vertical column of 3D space.
The top-down images simplify reasoning with pick-and-place operations, as coordinates on the workspace can be directly mapped to the observation pixel space, and improving training data augmentation process \cite{platt2019deictic}, as translating and rotating the observation can quickly generate different configurations of the object.
Thus, both Transporter frameworks have shown significant success in pick-and-place tasks, including stacking, sweeping, and rearranging deformable objects \cite{zeng2020transporter,seita_bags_2021}.

GCTN consists of 4 Fully Convolutional Networks (FCNs): one FCN is used for picking module and three FCNs for placing module.
All FCNs are hourglass encoder-decoder residual networks (ResNets) \cite{he2016deep} with 43-layers and 9.9M parameters, made up of 12 residual blocks and 8-strides. 

\subsubsection{Picking Module} 
The picking module consists of one FCN, $f_\text{pick}$.
It takes in as input the stacked image of the observation $\mbo_t$ and goal $\mbo_g$ images, and outputs dense pixel-wise prediction of action-values $\mathcal{Q}_\text{pick}$ that correlate with inferred picking success.
This gives us the goal-conditioned picking action $\mathcal{T}_\text{pick} = \argmax_{(u,v)}\{\mathcal{Q}_\text{pick}((u,v)|\mbo_t, \mbo_g)\}$, with $(u,v)$ as the pixel corresponding to the picking action determined by the camera-to-robot vision mapping.

\subsubsection{Placing Module} 
The placing module consists of three FCNs, $\Phi_\text{key}, \Phi_\text{query}, \Phi_\text{goal}$.
First, $\Phi_\text{key}$ and $\Phi_\text{query}$ take in the observation $\mbo_t$ to output dense pixel-wise feature maps of the inputs.
Then, $\Phi_\text{goal}$ takes in the goal image $\mbo_g$ to also output a pixel-wise feature map, which is then multiplied to each of the outputs of $\Phi_\text{key}$ and $\Phi_\text{query}$ using the Hadamard product to produce pixel-wise correlated features $\psi$:
\begin{align}
    \psi_\text{query}(\mbo_t, \mbo_g) &=  \Phi_\text{query}(\mbo_t) \odot \Phi_\text{goal}(\mbo_g),\\
    \psi_\text{key}(\mbo_t, \mbo_g) &=  \Phi_\text{key}(\mbo_t) \odot \Phi_\text{goal}(\mbo_g).
\end{align}
Finally, the query feature is cropped around the pick action $\psi_\text{query}(\mbo_t, \mbo_g)[\mathcal{T}_\text{pick}]$, then cross-correlated with the key feature to produce dense pixel-wise prediction of action-values $\mathcal{Q}_\text{place}$ that correlate with inferred placing success:
\begin{align}
    \mathcal{Q}_\text{place}&(\Delta \tau|\mbo_t, \mbo_g, \mathcal{T}_\text{pick}) \notag\\
    &= \psi_\text{query}(\mbo_t, \mbo_g)[\mathcal{T}_\text{pick}]\; * \;\psi_\text{key}(\mbo_t, \mbo_g)[\Delta \tau].
\end{align}
The convolution covers the space of all possible placing poses through $\Delta \tau$, and gives us the goal-conditioned placing action $\mathcal{T}_\text{place} = \argmax_{\Delta \tau}\{\mathcal{Q}_\text{place}(\Delta \tau|\mbo_t, \mbo_g, \mathcal{T}_\text{pick})\}$. 

\begin{figure*}[t]
\centering
\vspace{0.0625in}
  \includegraphics[width=\textwidth]{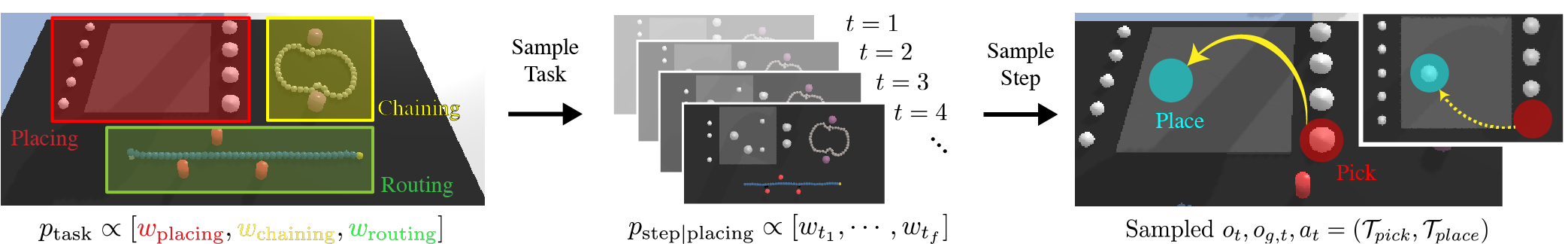}
  \caption{\textbf{Multi-Task Weighted Sampling Training.} For a selected demo, we first choose which task to sample from with probability $p_\text{task}$, then choose which step to sample from within the chosen task with probability $p_\text{step}$, getting $(\mbo_t, \mba_t, \mbo_{g,t})$.}
  \vspace{-1.5em}
  \label{fig:sampling}
\end{figure*}
\section{System Overview}
\label{sec:system}
We present our two main architecture contributions: (1) Sequence-Conditioned Transporter Networks (SCTN), which can effectively leverage structure and compositionality of multi-task problems through demonstration image sequences; and (2) multi-task learning with weighted sampling prior that improves learning robustness and efficiency.

\subsection{Sequence-Conditioned Transporter Networks}
\label{subsec:sctn}
We first determine what form of intermediate representation is suitable for \textit{demonstration sequences}. 
One way to represent sequential information is demonstration sequence images \cite{dasari2020transformer,yu2018oneshot}. 
We collect demonstration sequence images at every time step $t$ we receive a positive delta reward: whenever the expert demonstrator successfully places an object at a desired location or completes a part of a sequential task, we record the resulting observation.
Thus, after a successful expert demonstration with $D$ delta reward signals, we will have collected demonstration sequence images $\mbv \equiv \{\mbv_i\}_{i=1}^{D}$ that reaches a desired final goal configuration.

Next, we introduce the \textit{next-step goal module}, a general framework for determining the next suitable goal given an observation $\mbo_t$.
The next-step goal module seeks to find the closest delta reward time step in the demonstration image sequence, and infer the next-step goal as the demonstration image after the matched step.
There are many possible implementations of this module: it could be a raw image distance-based module that compares the distance between the current observation $\mbo_t$ against the demonstration sequence images $\mbv_i$, or a neural network-based module that learns and compares temporal embeddings of $\mbv_i,\mbo_t$ \cite{Sermanet2017TCN}.

In this work, we set the next-step goal module as an oracle operator that knows which delta reward time step the agent is in and outputs the next-step goal image from $\mbv$ during evaluation, for ease of implementation and to demonstrate proof of concept.
This oracle operator is reasonable since in our real robot system, a human operator checks each step of the robot action and gives signal to continue or stop.

Thus, SCTN takes in as input the observation image $\mbo_t$ and the demonstration sequence $\mbv$ of the task being solved into a desired final configuration, determines the next-step goal image $\mbo_{g,t}$ from $\mbv$, and passes $(\mbo_t, \mbo_{g,t})$ into GCTN to obtain the pick-and-place action $\mba_t$ (\cref{fig:sctn}).
We note that this is a general system architecture that could potentially work well with any discrete-time goal-conditioned agents.

\subsection{Multi-Task Training with Weighted Sampling}
\label{subsec:multitask}
We introduce next-step goal-conditioning, as well as task-level and step-level weighting that serve as useful sampling priors for stochastic gradient steps in training SCTN.

\subsubsection{Next-Step Goal-Conditioning}
To train the next-step goal-conditioning architecture, we sample $(\mbo_t, \mba_t, \mbo_{g,t})$ from a given demonstration episode, with $\mbo_{g,t}$ as the next-step goal image within the episode. 
This approach is based on Hindsight Experience Relay \cite{andrychowicz2018hindsight}, where the goal image $\mbo_{g,t}$ is the first state with a delta reward signal after $\mbo_t$.

\subsubsection{Weighted Sampling for Multi-Task Problems}
To train SCTN, a demonstration $\zeta_i$ is randomly chosen from a dataset of $N$ demonstrations $\mathcal{D} = \{\zeta_1, \zeta_2, \cdots, \zeta_N\}$, after which a step $t$ of the demo needs to be sampled from the set of all time steps $t \in [1, \cdots, T_i]$ to give us the training sample $(\mbo_t, \mba_t, \mbo_{g,t})$.
However, sampling uniformly across time steps can be problematic for multi-task problems that consist of tasks with varying difficulties and horizons.

We introduce weighted sampling on both task-level and step-level to leverage task structure information.
For task-level weighting, we assign each task $i$ in a multi-task problem a weight $w_i$.
We use two weighting schemes: (i) uniform weighting (``Unif'') that gives all tasks equal weights, and (ii) complexity weighting (``Comp'') that weighs a given task proportional to its inferred task complexity.
The inferred task complexity of a task is how many gradient steps it takes to reach optimal evaluation performance.
Then for step-level weighting, we assign each step $t$ of the demonstration corresponding for each task a weight $w_{i,t}$.
We assume that the expert demonstration solves each task sequentially, which means there is no mixing of demo steps between tasks.

To determine the inferred task complexity and the step-level weighting for each task, we employ black-box optimization \cite{vizier} to obtain the values in \cref{table:weighting}.
With the weights determined, we can now enhance multi-task learning with weighted sampling.
We first choose a task $i$ with probability $p_\text{task}\propto \{w_i\}, i\in \{\text{tasks}\}$, then choose a step $t$ with probability $p_{\text{step}|i}\propto \{w_{i,t}\}, t\in [t_{i, 1}, \cdots, t_{i, f}]$ where $t_{i, 1}$ and $t_{i,f}$ denote the first and last steps of the demo episode that pertains to solving task $i$.
\cref{fig:sampling} depicts the weighted sampling process for training SCTN.

\begin{table*}[t]
  \parbox{.55\linewidth}{
\vspace{0.0625in}
      \setlength\tabcolsep{1.2pt}
      \centering
      \begin{tabular}{lrrrrrr}
      \toprule
      & Precise & Sequential & Occluded & Deformable & Lengthy & Goal Objects $n$\\
      Module & Placing & Planning & Goal & Objects & Task (6+) &(Max. Steps)\\
      \midrule
      \textit{placing}   & \cmark & \xmark & \xmark & \xmark & \cmark & 9 $(n+2)$\\
      \textit{stacking}   & \cmark & \cmark & \cmark & \xmark & \cmark & 6 $(n+2)$\\
      \textit{chaining}   & \xmark & \xmark & \xmark & \cmark & \xmark & 2 $(n+2)$\\
      \textit{routing}   & \xmark & \cmark & \xmark & \cmark & \xmark & 2-3 $(n+4)$\\
      \bottomrule
      \end{tabular}
  }
  \parbox{.37\linewidth}{
\vspace{0.0625in}
      \setlength\tabcolsep{4.0pt}
      \centering
      \begin{tabular}{lll}
      \toprule
      Single-Task & Double-Task & Triple-Task\\
      Problems & Problems & Problems\\
      \midrule
      1. \textit{placing} & 5. \textit{placing + routing} & 9. \textit{placing + chaining}\\
      2. \textit{stacking} & 6. \textit{stacking + routing} &  \;\;\;\;\;+ \textit{routing}\\
      3. \textit{chaining} & 7. \textit{placing + chaining} & 10. \textit{placing + stacking}\\
      4. \textit{routing} & 8. \textit{placing + stacking} &  \;\;\;\;\;\;+ \textit{routing}\\
      \bottomrule
      \end{tabular}
  }
  \caption{\textbf{MultiRavens}. Overview of task modules and their attributes (left), and the 10 multi-task problems we test in this work (right).}
  \vspace{-1.5em}
  \vspace{-0.0625in}
  \label{table:task-attributes}
\end{table*}
\section{Simulator and Tasks}
\label{sec:tasks}
We evaluate the system on \textit{MultiRavens}, a novel suite of simulated compositional manipulation tasks built on top of PyBullet \cite{coumans2016pybullet} with an OpenAI gym \cite{brockman2016openai} interface. 

\subsection{Modularized Tasks in PyBullet}
\label{sec:pybullet}
To develop standardized benchmarks for compositional tasks, we propose a new PyBullet simulation platform, \textit{MultiRavens}, based on Ravens \cite{zeng2020transporter} and DeformableRavens \cite{seita_bags_2021}, which allows support for loading multiple tasks at once.
PyBullet is a widely-used publicly available simulator for robotics research, and the prior Ravens frameworks support various discrete-time tabletop manipulation tasks.
In MultiRavens, we introduce three new task modules: \textit{placing, chaining} and \textit{routing}; as well as modify one of the existing tasks from Ravens: \textit{stacking}.
The tasks take inspiration from the National Institute of Standards and Technology (NIST) Assembly Task Boards \cite{iros2020challenge}.
MultiRavens fully embraces compositionality of tasks by allowing full customizability of task modules in compositional tasks.

Even though the PyBullet implementations of these tasks are simplified, the four task modules capture the compositionality of industrial tasks while increasing the difficulty of pick-and-place reasoning compared to Ravens and DeformableRavens.
The challenges include precisely placing without direct visual cues for long horizon, dealing with deformable objects with other static objects present, and planning around occluded goals without memorizing task sequences.
\cref{fig:tasks} shows the four task modules and an example multi-task problem, and \cref{table:task-attributes} provides a quick description for each \textit{placing, routing, chaining} and \textit{stacking}. 

\subsection{Benchmark for Multi-Task Pick-and-Place Problems}
\label{sec:benchmark}
We benchmark our agents on 10 discrete-time multi-task tabletop manipulation problems from MultiRavens in \cref{table:task-attributes}.
Each environment setup consists of a Universal Robot arm UR5 with a suction gripper, a $0.5\times 1$m tabletop workspace where task modules are loaded, and 3 calibrated RGB-D cameras with known intrinsics and extrinsics, diagonally overlooking the workspace.
The three RGB-D images are augmented to produce a single top-down observation $\mbo_t \in \mathbb{R}^{320\times160\times 6}$, which has pixel resolution of $320\times160$ representing a $3.125\times3.125$mm vertical column of 3D space with 3 RGB channel and 3 channel-wise depth values.
In principle, this camera setup can be replaced with a single top-down camera, but the image reconstruction method gives us a clear view of the workspace without the UR5 arm.

We use the same grasping and releasing motion primitives in the Transporter works \cite{zeng2020transporter,seita_bags_2021}, which locks in the degrees of freedom of the end effector and the object using fixed constraints to pick and lift an object up.
Furthermore, we use different motion primitive settings for different tasks, determined by which tasks we have completed up until a given step.
For instance, for \textit{routing}, we use a primitive setting that makes the arm move slowly at a low lifting height to reduce unnecessary cable movements and prevent undoing any previous successful routes.
For \textit{stacking}, we use a setting that lets the arm move quickly at a higher height to be able to stack blocks on top of each other.

\section{Experiments}
\label{sec:experiments}
We use stochastic demonstrator policies to obtain $N=1000$ demonstration data per problem.
Each policy we benchmark below are then trained with 1, 10, 100 or 1000 demonstrations.
For each problem, we test GCTN and SCTN, both with various weighted sampling schemes to analyze whether the system architecture of sequence-conditioning and weighted sampling helps with the performance.
Training details, such as learning rate, optimizer, and data augmentation, remain the same as Transporter works \cite{zeng2020transporter,seita_bags_2021}.

For single-task problems, we test the following four agents: GCTN, GCTN-Step, SCTN, and SCTN-Step. 
GCTN-Step and SCTN-Step are Transporter agents that are trained with step-level weighting for a given single-task problem.
For multi-task problems, we test the following six agents: GCTN, GCTN-Unif-Step, GCTN-Comp-Step, SCTN, SCTN-Unif-Step, SCTN-Comp-Step. 
The agents with Unif-Step suffixes are trained with uniform task-level weighting and step-level weighting, and those with Comp-Step are trained with inferred task complexity task-level weighting and step-level weighting.
The weighting schemes are given in \cref{table:weighting}.

Zeng et al. \cite{zeng2020transporter} and Seita et al. \cite{seita_bags_2021} also test \textit{GT-State MLP} models in their analyses, which use ground truth pose information of each objects in the scene and process the input with multi-layer perception (MLP) networks.
However, two of our task modules, \textit{chaining} and \textit{routing}, have variable number of objects in the scene due to randomized chain and cable lengths.
This renders the na\"ive ground truth-based models unapplicable to these problems.
Other directly comparable vision-based discrete-time pick-and-place agents tested in Zeng et al. \cite{zeng2020transporter}, such as \textit{Form2Fit} \cite{zakka2019form2fit} and \textit{ConvMLP} \cite{levine2016end}, may potentially be made comparable to GCTN and SCTN.
However, these agents are not goal-conditioned.

While it is possible to substantially modify these agents, these models are shown to perform rather poorly compared to the Transporter variants and will require nontrivial modifications to perform reasonably well on these tasks.
Since we are mainly interested in analyzing the benefits of the new systems architecture of multi-task learning and sequence-conditioning, we focus our analysis on how different weighting and conditioning schemes affect the performance.

\begin{table*}[t]
\centering
    \parbox{.37\linewidth}{
\vspace{0.0625in}
      \setlength\tabcolsep{2.3pt}
      \centering
      \scriptsize
      \begin{tabular}{@{}lrrrr}
      \toprule
       & Inferred & Weighting Scheme & Mean & Mean \\
      Module & Complexity & [min. \& max. weights]& Completion & Reward\\
      \midrule
      \textit{placing}   & 4K Steps & Linear: [1.0 \linweight 2.76] & 100\% & 1.0\\
      \textit{chaining}   & 20K Steps & Last: [1.0 \lastweight 1.92] & 90\% & 0.95\\
      \textit{routing}   & 20K Steps & Linear: [1.0 \linweight 1.76] & 25\% & 0.52\\
      \textit{stacking}   & 12K Steps & Linear: [1.0 \linweight 1.50] & 90\% & 0.94\\
      \bottomrule
      \end{tabular}
      \vspace{0.25em}
      \caption{Hyperparameters for weighted sampling and evaluation statistics for each task.}
      \vfill
      
  \vspace{-0.0625in}
      \label{table:weighting}
  }
  \parbox{0.02\linewidth}{\hfill\;}
  \parbox{.57\linewidth}{
\vspace{0.0625in}
      \setlength\tabcolsep{1.5pt}
      \centering
      \scriptsize
      \begin{tabular}{@{}laaaabbbbaaaabbbbr@{}}
      \toprule
       & \multicolumn{4}{c}{1. \textit{placing}}  & \multicolumn{4}{c}{2. \textit{stacking}}  & \multicolumn{4}{c}{3. \textit{chaining}}  & \multicolumn{4}{c}{4. \textit{routing}}  \\
        \cmidrule(l{3pt}r{2pt}){2-5} \cmidrule(l{3pt}r{2pt}){6-9} \cmidrule(l{3pt}r{2pt}){10-13} \cmidrule(l{3pt}r{2pt}){14-17}
        Method & 1 & 10 & 100 & 1000 & 1 & 10 & 100 & 1000 & 1 & 10 & 100 & 1000 & 1 & 10 & 100 & 1000 \\
        \midrule
        GCTN  & 37.0 & 44.0 & 39.0 & 37.0 & 0.0 & 0.0 & 0.0 & 0.0 & 4.0 & 17.0 & 19.0 & 2.0 & \textbf{13.0} & \textbf{66.0} & 69.0 & \textbf{75.0} \\
        GCTN-Step  & 20.0 & 35.0 & 29.0 & 37.0 & 0.0 & 0.0 & 0.0 & 0.0 & 4.0 & 22.0 & 1.0 & 8.0 & 13.0 & 64.0 & \textbf{78.0} & 68.0 \\
        \textbf{SCTN}  & \textbf{94.0} & \textbf{100.0} & \textbf{100.0} & \textbf{100.0} & 0.0 & 47.0 & 53.0 & \textbf{70.0} & \textbf{10.0} & 26.0 & \textbf{36.0} & 27.0 & 1.0 & 10.0 & 12.0 & 15.0 \\
        \textbf{SCTN-Step}  & 77.0 & 100.0 & 98.0 & 80.0 & \textbf{3.0} & \textbf{50.0} & \textbf{73.0} & 67.0 & 7.0 & \textbf{30.0} & 28.0 & \textbf{33.0} & 1.0 & 11.0 & 10.0 & 10.0 \\
      \bottomrule
      \end{tabular}
      \caption{\textbf{Single-task results.} Completion rate vs. number of demonstration episodes used in training.}
      \vspace{-1em}
      \label{table:single}
     }
     \parbox{0.01\linewidth}{\hfill\;}
\end{table*}
\begin{table*}[t]
  \setlength\tabcolsep{2.5pt}
  \centering
  \scriptsize
  \begin{tabular}{@{}laaaabbbbaaaarr@{}}
  \toprule
     & \multicolumn{4}{c}{5. \textit{placing + routing}}  & \multicolumn{4}{c}{6. \textit{stacking + routing}}  & \multicolumn{4}{c}{7. \textit{placing + chaining}}  \\
    \cmidrule(l{3pt}r{2pt}){2-5} \cmidrule(l{3pt}r{2pt}){6-9} \cmidrule(l{3pt}r{2pt}){10-13}
    Method & 1 & 10 & 100 & 1000 & 1 & 10 & 100 & 1000 & 1 & 10 & 100 & 1000 \\
    \midrule
    GCTN  & 1.0 {\color{gray}(4.8)} & 28.0 {\color{gray}(29.0)} & 20.0 {\color{gray}(26.9)} & 25.0 {\color{gray}(27.7)} & 0.0 {\color{gray}(0.0)} & 0.0 {\color{gray}(0.0)} & 0.0 {\color{gray}(0.0)} & 0.0 {\color{gray}(0.0)} & 3.0 {\color{gray}(1.5)} & 4.0 {\color{gray}(7.5)} & 9.0 {\color{gray}(7.4)} & 2.0 {\color{gray}(0.7)} \\
    GCTN-Unif-Step  & 3.0 {\color{gray}(2.6)} & 25.0 {\color{gray}(22.4)} & 28.0 {\color{gray}(22.6)} & 29.0 {\color{gray}(25.2)} & 0.0 {\color{gray}(0.0)} & 0.0 {\color{gray}(0.0)} & 0.0 {\color{gray}(0.0)} & 0.0 {\color{gray}(0.0)} & 1.0 {\color{gray}(0.8)} & 7.0 {\color{gray}(7.7)} & 6.0 {\color{gray}(0.3)} & 8.0 {\color{gray}(3.0)} \\
    GCTN-Comp-Step  & 1.0 {\color{gray}(2.6)} & 5.0 {\color{gray}(22.4)} & 18.0 {\color{gray}(22.6)} & 13.0 {\color{gray}(25.2)} & 0.0 {\color{gray}(0.0)} & 0.0 {\color{gray}(0.0)} & 0.0 {\color{gray}(0.0)} & 0.0 {\color{gray}(0.0)} & 2.0 {\color{gray}(0.8)} & 2.0 {\color{gray}(7.7)} & 0.0 {\color{gray}(0.3)} & 1.0 {\color{gray}(3.0)} \\
    SCTN  & \textbf{5.0 {\color{gray}(0.9)}} & 46.0$^\dagger$ {\color{gray}(10.0)} & 62.0$^\dagger$ {\color{gray}(12.0)} & 50.0$^\dagger$ {\color{gray}(15.0)} & 0.0 {\color{gray}(0.0)} & \textbf{35.0$^\dagger$ {\color{gray}(4.7)}} & \textbf{54.0$^\dagger$ {\color{gray}(6.4)}} & 39.0$^\dagger$ {\color{gray}(10.5)} & \textbf{4.0 {\color{gray}(9.4)}} & 33.0 {\color{gray}(26.0)} & 40.0 {\color{gray}(36.0)} & 44.0$^\dagger$ {\color{gray}(27.0)} \\
    \textbf{SCTN-Unif-Step}  & 4.0 {\color{gray}(0.8)} & \textbf{50.0$^\dagger$ {\color{gray}(11.0)}} & \textbf{81.0$^\dagger$ {\color{gray}(9.8)}} & \textbf{86.0$^\dagger$ {\color{gray}(8.0)}} & 0.0 {\color{gray}(0.0)} & 25.0$^\dagger$ {\color{gray}(5.5)} & 45.0$^\dagger$ {\color{gray}(7.3)} & \textbf{65.0$^\dagger$ {\color{gray}(6.7)}} & 3.0 {\color{gray}(5.4)} & \textbf{38.0 {\color{gray}(30.0)}} & \textbf{48.0$^\dagger$ {\color{gray}(27.4)}} & \textbf{50.0$^\dagger$ {\color{gray}(26.4)}} \\
    \textbf{SCTN-Comp-Step}  & 2.0 {\color{gray}(0.8)} & 25.0$^\dagger$ {\color{gray}(11.0)} & 38.0$^\dagger$ {\color{gray}(9.8)} & 44.0$^\dagger$ {\color{gray}(8.0)} & 0.0 {\color{gray}(0.0)} & 30.0$^\dagger$ {\color{gray}(5.5)} & 49.0$^\dagger$ {\color{gray}(7.3)} & 49.0$^\dagger$ {\color{gray}(6.7)} & 1.0 {\color{gray}(5.4)} & 16.0 {\color{gray}(30.0)} & 46.0$^\dagger$ {\color{gray}(27.4)} & 32.0 {\color{gray}(26.4)} \\
    \midrule & \multicolumn{4}{c}{8. \textit{placing + stacking}}  & \multicolumn{4}{c}{9. \textit{placing + chaining + routing}}  & \multicolumn{4}{c}{10. \textit{placing + stacking + routing}}  \\
    \cmidrule(l{3pt}r{2pt}){2-5} \cmidrule(l{3pt}r{2pt}){6-9} \cmidrule(l{3pt}r{2pt}){10-13}
    Method & 1 & 10 & 100 & 1000 & 1 & 10 & 100 & 1000 & 1 & 10 & 100 & 1000 \\
    \midrule
    GCTN  & 0.0 {\color{gray}(0.0)} & 0.0 {\color{gray}(0.0)} & 0.0 {\color{gray}(0.0)} & 0.0 {\color{gray}(0.0)} & 1.0 {\color{gray}(0.2)} & 5.0 {\color{gray}(4.9)} & 4.0 {\color{gray}(5.1)} & 9.0 {\color{gray}(0.6)} & 0.0 {\color{gray}(0.0)} & 0.0 {\color{gray}(0.0)} & 0.0 {\color{gray}(0.0)} & 0.0 {\color{gray}(0.0)} \\
    GCTN-Unif-Step  & 0.0 {\color{gray}(0.0)} & 0.0 {\color{gray}(0.0)} & 0.0 {\color{gray}(0.0)} & 0.0 {\color{gray}(0.0)} & 0.0 {\color{gray}(0.1)} & 4.0 {\color{gray}(4.9)} & 9.0 {\color{gray}(0.2)} & 8.0 {\color{gray}(2.0)} & 0.0 {\color{gray}(0.0)} & 0.0 {\color{gray}(0.0)} & 0.0 {\color{gray}(0.0)} & 0.0 {\color{gray}(0.0)} \\
    GCTN-Comp-Step  & 0.0 {\color{gray}(0.0)} & 0.0 {\color{gray}(0.0)} & 0.0 {\color{gray}(0.0)} & 0.0 {\color{gray}(0.0)} & 0.0 {\color{gray}(0.1)} & 2.0 {\color{gray}(4.9)} & 1.0 {\color{gray}(0.2)} & 4.0 {\color{gray}(2.0)} & 0.0 {\color{gray}(0.0)} & 0.0 {\color{gray}(0.0)} & 0.0 {\color{gray}(0.0)} & 0.0 {\color{gray}(0.0)} \\
    SCTN  & \textbf{1.0 {\color{gray}(0.0)}} & 44.0 {\color{gray}(47.0)} & 54.0 {\color{gray}(53.0)} & 56.0 {\color{gray}(70.0)} & 1.0 {\color{gray}(0.1)} & \textbf{27.0$^\dagger$ {\color{gray}(2.6)}} & 33.0$^\dagger$ {\color{gray}(4.3)} & 27.0$^\dagger$ {\color{gray}(4.0)} & 0.0 {\color{gray}(0.0)} & 21.0$^\dagger$ {\color{gray}(4.7)} & 35.0$^\dagger$ {\color{gray}(6.4)} & 41.2$^\dagger$ {\color{gray}(10.5)} \\
    \textbf{SCTN-Unif-Step}  & 1.0 {\color{gray}(2.3)} & 57.0 {\color{gray}(50.0)} & \textbf{68.0 {\color{gray}(71.5)}} & 51.0 {\color{gray}(53.6)} & \textbf{2.0 {\color{gray}(0.1)}} & 26.0$^\dagger$ {\color{gray}(3.3)} & \textbf{49.0$^\dagger$ {\color{gray}(2.7)}} & \textbf{53.8$^\dagger$ {\color{gray}(2.6)}} & 0.0 {\color{gray}(0.0)} & \textbf{42.5$^\dagger$ {\color{gray}(5.5)}} & 44.0$^\dagger$ {\color{gray}(7.2)} & \textbf{48.7$^\dagger$ {\color{gray}(5.4)}} \\
    \textbf{SCTN-Comp-Step}  & 0.0 {\color{gray}(2.3)} & \textbf{58.0 {\color{gray}(50.0)}} & 64.0 {\color{gray}(71.5)} & \textbf{62.0 {\color{gray}(53.6)}} & 0.0 {\color{gray}(0.1)} & 21.0$^\dagger$ {\color{gray}(3.3)} & 32.0$^\dagger$ {\color{gray}(2.7)} & 39.0$^\dagger$ {\color{gray}(2.6)} & 0.0 {\color{gray}(0.0)} & 23.0$^\dagger$ {\color{gray}(5.5)} & \textbf{49.0$^\dagger$ {\color{gray}(7.2)}} & 29.0$^\dagger$ {\color{gray}(5.4)} \\
  \bottomrule
  \end{tabular}
  \caption{\textbf{Multi-task results.} Completion rate vs. number of demonstration episodes used in training. For both Tables \ref{table:single} and \ref{table:multi}, the completion rate is calculated as mean \% over $20\times5$ test-time episodes in simulation of the best saved snapshot, the \# of demos used in training are chosen from $\{1, 10, 100, 1000\}$, and bold indicates the best performance for a given problem and \# of demos. For \cref{table:multi}, gray number in parentheses indicates expected performance of combining single-task Transporter agents, and dagger $(\dagger)$ indicates more than 10\% completion rate improvement compared to the expected performance.}
  \vspace{-2em}
  \label{table:multi}
\end{table*}
\section{Results}
\label{sec:results}
We summarize results in Tables \ref{table:single} and \ref{table:multi}. 
For each agent and number of demo episodes $N \in \{1,10,100,1000\}$, we train the models 5 times with different TensorFlow \cite{tensorflow} random seeds, trained for $20\text{K}\times m$ iterations with batch size of 1, where $m$ is the number of task modules in a given problem.
We evaluate the trained model every tenth of the training iterations, where the trained model snapshot is validated on a roll out of 20 evaluation episodes.
This results in $20 \times 5 =100$ metrics for 10 snapshots of the model. 
We evaluate agent performances based on completion rate, which aptly captures agents' abilities to reliably complete all tasks in a given episode.
For completion rate, we assign the value of 1 for a given episode if and only if all the task modules are completed, and otherwise 0.
We average the completion rates over each snapshot and report the maximum over all snapshots in Tables \ref{table:single} and \ref{table:multi}.

\subsection{Single-Task}
\label{sec:single-tasks}
For single-task problems, both variants of SCTNs outperform GCTNs in \textit{placing} and \textit{stacking}.
SCTN has a remarkably high completion rate of 94\% for \textit{placing} even with just 1 demo episode, while GCTNs hover around 30-40\%.
GCTNs often suffer from two types of errors: inaccurately placing the disks, and stacking a disk on top of another one.
In these scenarios, GCTN does not know how to properly correct itself since the top-down view is unable to capture the stacking mistake and the attention module mostly focuses only on the disks that are directly outside the square plate.

SCTN-Step also has a high completion rate of 50\% with 10 demos and 73\% with 100 demos for \textit{stacking}, while GCTNs completely fail, which is expected as GCTNs cannot reason with blocks that are (partially) occluded from the top-down view.
Interestingly, GCTNs successfully imitate the overall behavior of stacking the blocks but completely disregard the goal configuration (\cref{fig:failure}).

All Transporter agents perform similarly poorly on the \textit{chaining} task, while GCTNs perform far better on \textit{routing} with completion rate of 66\% with only 10 demos.
SCTNs often struggle at performing the final cable stretching move to match the goal configuration and seem to prematurely consider that it has already completed the task, then perform out-of-distribution moves that lead to failure.
This seems to be an artifact of both overfitting and task complexity, as efficient routing demonstration steps can overshadow the complexities of dealing with linear deformable objects.

Thus, for single-task problems, SCTNs perform far better and more efficiently than GCTNs on tasks that deal with small individual objects over long horizons, but does not perform better than GCTNs on tasks that deal with deformable objects.
On a single-task level, the step-level weighting does not seem to improve the performance significantly.

\subsection{Multi-Task}
\label{sec:multi-tasks}
For all multi-task problems, SCTNs vastly outperform GCTNs.
We also report the \textit{expected completion rate} in \cref{table:multi}, which is the completion rate we can expect if we use separate Transporter agents of same settings for each task in the multi-task problem, calculated by multiplying the mean completion rates for each task.
SCTNs with weighted sampling often performed far better on multi-task problems than the expected completion rate.
For \textit{placing + routing}, SCTN-Unif-Step has a completion rate of 81\%, while the expected completion rate is only 9.8\%, which means SCTN performs better on the \textit{routing} task when jointly trained on \textit{placing} and \textit{routing} than when solely trained on \textit{routing}.

\begin{figure}[t]
\centering
  \includegraphics[width=0.48\textwidth]{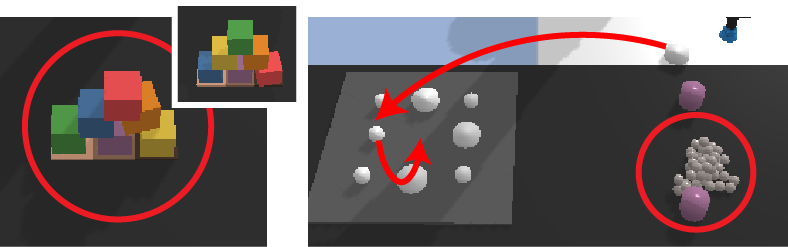}
  \caption{GCTN can imitate the block stacking behavior but mostly disregards the goal configuration (left), and makes several misplacing mistakes in multi-task problems (right).}
  \vspace{-1.5em}
  \label{fig:failure}
\end{figure}

This trend continues to show in triple-task problems, where SCTN-Unif-Step achieves 49.0\% completion rate with just 10 demos for \textit{placing + chaining + routing}.
Meanwhile, GCTN often makes several mistakes along the way (\cref{fig:failure}).
Thus, for multi-task problems, both sequence-conditioning and weighted sampling significantly improve the performance of the Transporter agents, often improving far beyond the expected performance when trained on multiple tasks.

\section{Conclusions and Future Works}
\label{sec:conclusions}
We introduce a suite of modularized tasks for benchmarking vision-based manipulation agents on multi-task problems.
In addition, we propose an architecture that enables discrete-time goal-conditioned pick-and-place agents to reason with sequence-conditioning and allows for more informed weighted sampling to tackle multi-task problems of long horizons.

Reasoning over multi-task problems is challenging, and prior works have shown mixed success results in multi-task training.
Our proposed system demonstrates high completion rates in many multi-task problems of long planning horizon when trained on as few as 10 demonstration episodes, and shows that sequence-conditioning with appropriate sampling schemes can bolster multi-task learning and planning.
Despite the promising results, it still remains a question how the new system architecture generalizes to other goal-conditioned agents.
Furthermore, while access to robots was limited due the COVID-19 pandemic, we aim to conduct physical experiments with suitable multi-task problems to see how well the system performs in realistic scenarios.

\section*{Acknowledgment}
This material is based upon work supported by Google-Berkeley AI Research Grant, DARPA Assured Autonomy Grant, and NSF GRFP (\# DGE 1752814).
Any opinions, findings, and conclusions or recommendations expressed in this material are those of the authors and do not necessarily reflect the views of any aforementioned organizations.
The authors also thank Gregory Lawrence, Katharina Muelling, Simon Stepputtis, Azarakhsh Keipour and Chelsea Finn for valuable discussions and manuscript feedback.

\clearpage
\bibliographystyle{IEEEtran}
\bibliography{IEEEabrv,references}

\clearpage

\setcounter{page}{1}
\setcounter{tocdepth}{2}

\startcontents
\noindent\textbf{Appendix Contents}
\vskip3pt\hrule\vskip5pt 
\printcontents{}{0}{} 
\vskip3pt\hrule\vskip5pt

\begin{appendices}
\section{Covid-19 Details}
\label{app:covid}
Due to the ongoing COVID-19 pandemic, performing physical robotics experiments became a huge challenge.
The resulting challenges include: restrictions in traveling to the robotic setup locations, restrictions in collecting demonstrations and performing experiments at the locations as per COVID-19 safety protocols, and difficulties and delays in sourcing materials to devise and build realistic yet tractable pick-and-place experiments.
On that end, we continue to utilize the PyBullet simulation environment like Zeng et al. \cite{zeng2020transporter} and Seita et al. \cite{seita_bags_2021}, and modify it to load multiple task modules at once.
We hope to investigate the performance of SCTN on realistic experiment setups in future works.

\section{Task Details}
\label{app:tasks}
\begin{figure}[t]
\centering
  \includegraphics[width=0.47\textwidth]{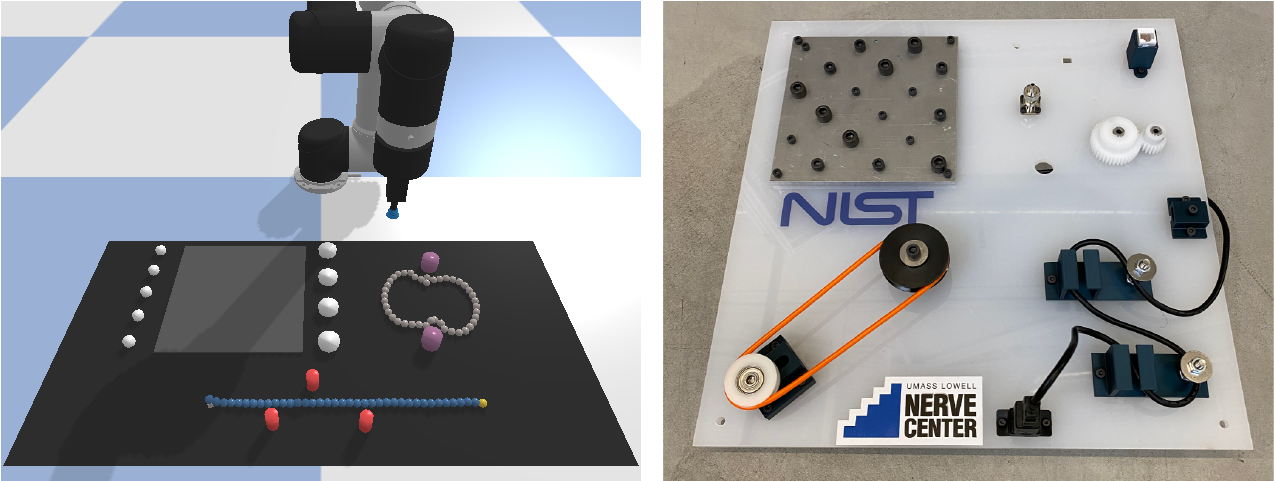}
  \caption{\small The new modularized tasks in MultiRavens, \textit{placing, chaining} and \textit{routing} (left), take inspiration from the National Institute of Standards and Technology (NIST) Assembly Task Boards (right).}
  \label{fig:nist-board}
  \vspace{-1.5em}
\end{figure}

In \textit{MultiRavens}, we propose 10 multi-task problems that are made up of 4 basic task modules.
These task modules take inspiration from National Institute of Standards and Technology (NIST) Assembly Task Boards \cite{iros2020challenge} as shown in \cref{fig:nist-board}.
In this section, we give further details of the task, including evaluation metrics and corresponding demonstrator policies.
The visualizations of these tasks are shown in \cref{fig:tasks}, and quick overview of unique task attributes of each task is given in \cref{table:task-attributes}.
All problems have task modules anchored at the same coordinates each time they are generated for consistency.

\subsection{Placing}
\subsubsection{Overview}
For the \textit{placing} task, the agent is tasked to place the 9 disks at their goal positions on top of the dark gray square plate. 
The 9 disks are initialized such that 5 of the ``even'' index disks are placed on the left of the plate, and 4 ``odd'' disks placed on the right.
The even and odd disks can either be big or small for a given parity, which gives us the total of 4 possible task goal configurations.

\subsubsection{Metrics}
The evaluation metric checks whether the disks are placed in their designated zone (even or odd) and whether they are placed close enough to the $3\times3$ grid of goal positions, where visualizations of green `+' corresponds to goal position for even disks and blue `+' for odd disks as shown in \cref{fig:goals}.
Delta reward of $1/9$ is given whenever a disk is placed successfully.
The metric is robust to stacking disks, which means that it does not award additional delta reward if there is already a disk at the goal position and another disk is stacked on top.
This ensures that the task is complete only when all the disks are in their designated goal configurations with high enough accuracy.

\subsubsection{Demonstrator Policy}
The even and odd disks have randomized oracle goal positions, which means that despite there being only 4 possible task goal configurations, the demonstration policy roll outs are highly varied and randomized.
The demonstrator randomly picks the disks to be placed at their designated oracle goal positions, which ensures that the Transporter agents cannot simply copy or memorize the sequences of moves.
The demonstrator often needs exactly 9 steps to successfully execute the demonstration.

\subsection{Chaining}
\subsubsection{Overview}
For the \textit{chaining} task, the agent is tasked to wrap the chain of beads around the two fixed pegs.
The peg distances and radii are randomly selected with 4 possible combinations, while the orientation of the task is set randomly with angle $\theta \in [-\pi/4, \pi/4]$.
The closed-loop chain is made up of connected beads in a manner similar to the implementations in the Transporter works \cite{zeng2020transporter,seita_bags_2021}.
Depending on the initial configuration, the bead chain length is variable, usually in the range of around 30-40 beads.

\subsubsection{Metrics}
The evaluation metric checks whether the chain is wrapped around the pegs, with each successful wrap resulting in a delta reward of $1/2$.
The evaluation goal poses check for whether there are any beads close to the points, shown as green `+' in \cref{fig:goals}.
The evaluation poses will only be fully satisfied if the chain is wrapped around both pegs without any twisting.

\subsubsection{Demonstrator Policy}
The demonstrator policy attempts to place the beads at each of their pre-specified bead-level goal poses shown as blue `+' in \cref{fig:goals}.
The demonstrator policy plans in two stages, where in each stage the demonstrator tries to place and wrap the chain over each peg.
Unlike other tasks, the \textit{chaining} task often requires more than 2 steps to successfully execute the demonstration, as the demonstration policy may sometimes need a couple of steps to properly place the chain around each peg.
However, it usually suffices to let the demonstrator policy plan for 3 steps in the demonstration roll outs for a reasonable demonstration success rate.

\subsection{Routing}
\subsubsection{Overview}
For the \textit{routing} task, the agent is tasked to route the cable of beads around the fixed pegs.
The peg positions are generated at random, with the randomized parameters set to be alternating such that the cable needs to wind in different directions every peg.
With the current randomization settings, the environment is initialized with 2 or 3 pegs at various positions.
Similar to the \textit{chaining} task, the cable is made up of connected beads, where the left end of the cable is anchored to the workspace.
The cable length once again varies with the task initialization, but are programmed such that the cable is sufficiently long to route the cable fully around all the pegs with some slack.

\subsubsection{Metrics}
Similar to the \textit{chaining} task, the evaluation metric checks whether the cable is wrapped around the pegs, which are shown as green `+' in \cref{fig:goals}.
There is another evaluation goal position after the pegs, which checks whether the cable is fully stretched out after routing around the pegs.
This ensures that the cable is properly routed around the peg, which demonstrator policy sometimes fails to do for the last peg.
It also adds on an additional challenge for the agents to learn, as the resulting cable stretching move at the last step is a different type of maneuver compared to placing the cables over the pegs via Reidemeister-type moves.

Unlike other tasks, the \textit{routing} task requires that all the pegs be routed sequentially, where each sequentially correct moves will result in a delta reward of $1/(n+1)$ where $n$ is the number of pegs (we divide by $(n+1)$, as the task requires $n$ routing moves and 1 stretching move to be completed).
This means that routing the cable around first and third pegs will only result in a reward for first peg.
We enforce this sequential planning rule, as we noticed that out-of-order peg routing can often cause the \textit{routing} task to fail due to the cable often getting stuck in some configuration and the demonstrator and agents cannot effectively finish the task without undoing the routing moves.

\subsubsection{Demonstrator Policy}
\begin{figure}[t]
\centering
  \includegraphics[width=0.48\textwidth]{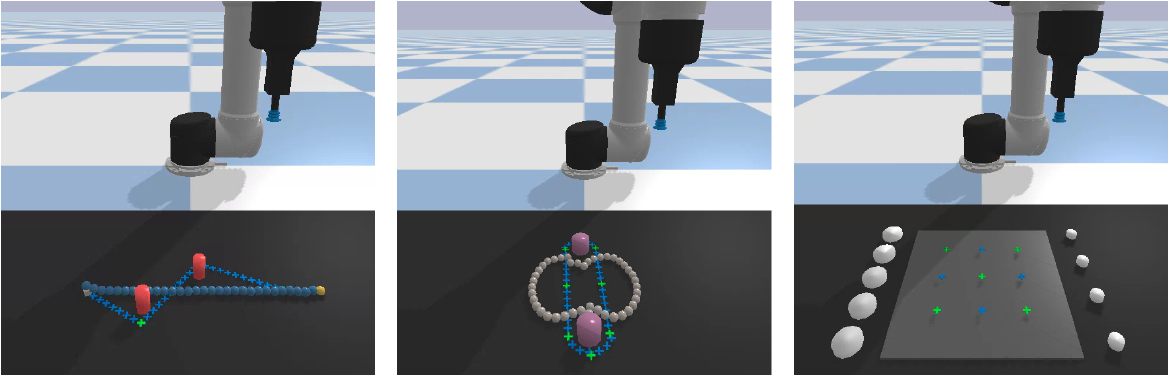}
  \caption{\small The explicitly visualized evaluation goal positions and/or object goal positions for each tasks: \textit{routing} (left), \textit{chaining} (middle) and \textit{placing} (right). For \textit{routing} and \textit{chaining}, the green `+' corresponds to the evaluation goal positions for reward calculation, and the blue `+' to the bead goal positions to aid oracle planning. For \textit{placing}, the green `+' corresponds to the goal positions for even disks, and the blue `+' for odd disks.}
  \label{fig:goals}
  \vspace{-1.5em}
\end{figure}

The demonstrator policy attempts to place the beads at each of their pre-specified bead-level goal poses shown as blue `+' in \cref{fig:goals}, similar to the \textit{chaining} task.
The demonstrator policy plans in multiple stages, where in each stage the demonstrator attempts to route the cable around the peg sequentially from left to right, using Reidemeister-type moves to lift the cable over the pegs and place it in the desired position.
For the last stage, the demonstrator then stretches the cable such that the last yellow-colored bead is in the desired goal position.

\subsection{Stacking}
\subsubsection{Overview}
For the \textit{stacking} task, it is similar to how it is defined in Ravens.
We make an additional modification that now the block goal positions are randomized with color such that goal-conditioning is necessary.
We also initialize the blocks in a $2\times3$ array for easy initialization and placement of task when loading multiple task modules together to avoid bad initialization that overlaps with other tasks.
The orientation for the base is randomized at 8 cardinal half-directions: $\theta \in \{\theta:\theta \equiv \pi k/ 4, k \in [0,\cdots, 7]\}$.

\subsubsection{Metrics}
The metrics are simple goal positioned metrics for each box, checking that each boxes are stacked in the right place within small margin of error.

\subsubsection{Demonstrator Policy}
The demonstrator policy operates similar to how it does in Ravens, where it stacks all three bottom boxes in the first stage, the next two middle boxes in the second stage, and the one top box in the third stage.

\subsection{Visualizations}
\begin{figure}[t]
\centering
  \includegraphics[width=0.48\textwidth]{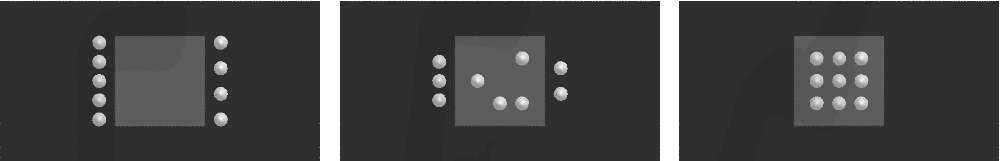}\\
  \vspace{0.3em}
  \includegraphics[width=0.48\textwidth]{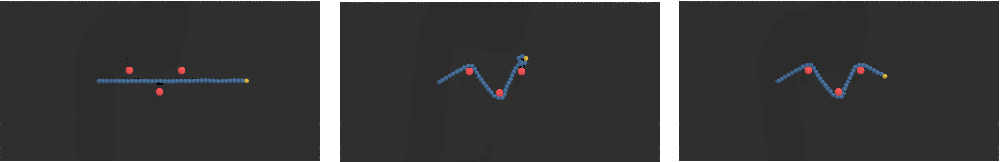}\\
  \vspace{0.3em}
  \includegraphics[width=0.48\textwidth]{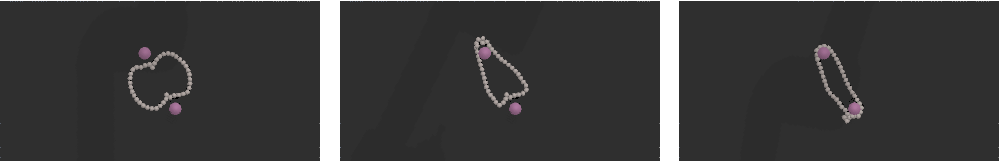}\\
  \vspace{0.3em}
  \includegraphics[width=0.48\textwidth]{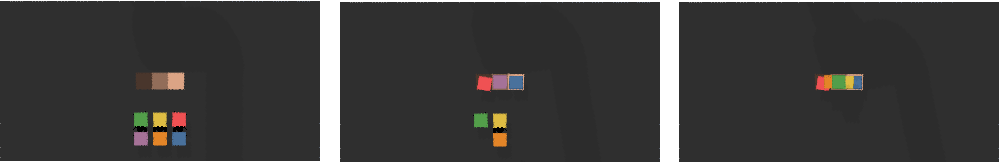}
  \caption{Visualization of the top-down image sequences of solving each task module: \textit{placing, routing, chaining}, and \textit{stacking} (top to bottom).}
  \label{fig:visualization}
  \vspace{-1.5em}
\end{figure}

In this section, we provide visualizations of the task initialization, intermediate states, and goal configurations for all 10 single and multi-task problems. 
For each problem, we provide sequences of 3 images, where the first image corresponds to the initial state, the middle image corresponds an arbitrary intermediate state, and the last image corresponds to the final goal configuration.
The complete set of visualizations is given in \cref{fig:visualization}.

\section{Determining Sampling Weights}
\label{app:hparam}
In \cref{subsec:multitask}, we have determined the task-level and step-level weights through black box optimization.
For the black box  optimization, we fix the number of demos to 1000, train the SCTN agent for 20K iterations, with snapshot evaluations taken every 2K iterations that is evaluated only on one random seed.
The evaluation metric for a given weighting scheme is the maximum evaluation performance on 20 test-time episodes over all 10 snapshots.

The hyperparameters to be optimized are the weighting scheme and the maximum weight $w_{\max} \in [1.5, 5.0]$.
The minimum weight is fixed to be 1 for all weighting schemes.
The three weighting schemes tested are the following:
\begin{itemize}
    \item \textbf{Linear:} Linearly increasing weighting scheme, where the weight of first step is 1.0 and the last step is $w_{\max}$, and the weights of other steps are interpolated linearly depending on the episode length.
    \item \textbf{Last:} Weighting scheme in which the weight of last step is $w_{\max}$ and all other steps are 1.0.
    \item \textbf{First-Last:} Weighting scheme in which the weight of first last steps are $w_{\max}$ and all other steps are 1.0.
\end{itemize}
We choose these weighting schemes since we often empirically observed that weighing the later steps of the demonstrations usually provided better learning for multi-task problems.

The inferred task complexity is determined by the first snapshot in which it achieves maximum evaluation performance. 

\section{Additional Experiment Results}
\label{app:results}
\begin{figure*}[t]
\centering
  \includegraphics[width=\textwidth]{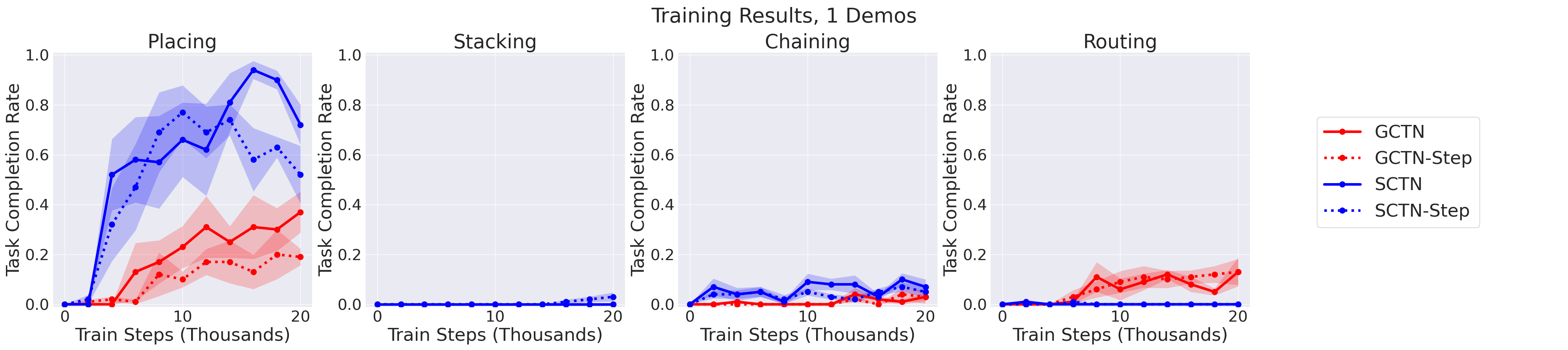}
  \includegraphics[width=\textwidth]{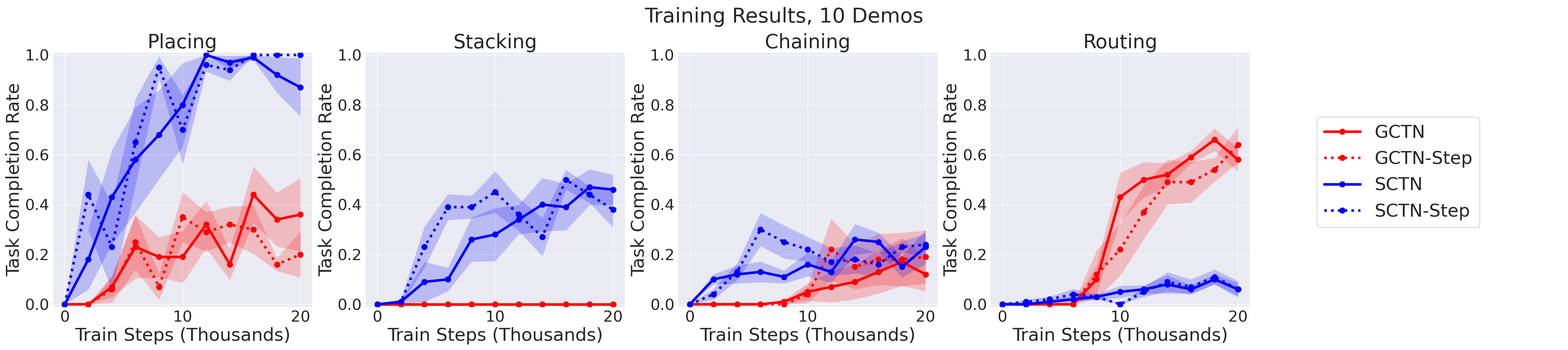}
  \includegraphics[width=\textwidth]{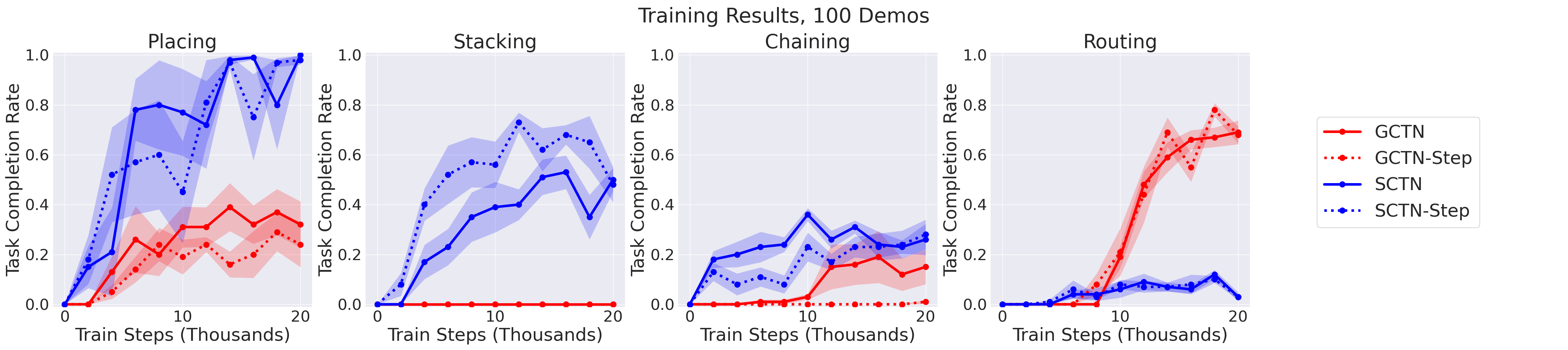}
  \includegraphics[width=\textwidth]{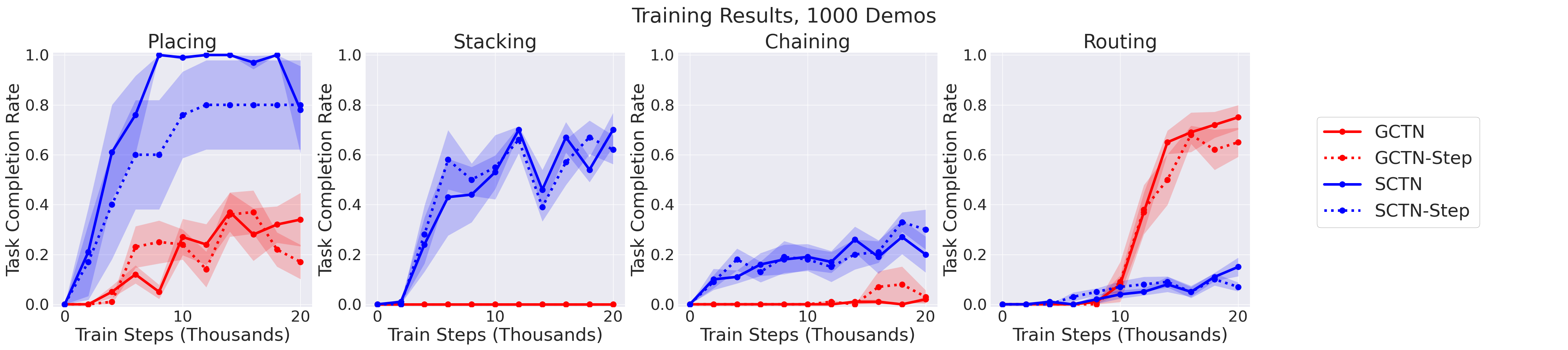}
  \caption{Results for models on the single-task problems, grouped by numbers of demonstrations the agents are trained on. The mean task completion rate over $20\times5$ test-time episode evaluations is plotted against the number of training steps in the unit of thousands. All single-task problems are trained with 20K training iterations.}
  \label{fig:single-demos}
  \vspace{-1.5em}
\end{figure*}

We provide more detailed evaluation statistics for the 10 single-task and multi-task problems.
\cref{fig:single-demos} contains learning curves showing the mean completion rates of different agents over $20\times5$ test-time episode evaluations as a function of training iterations for single-task problems, grouped by different numbers of demonstrations. 

\begin{figure*}[t]
\centering
  \includegraphics[width=\textwidth]{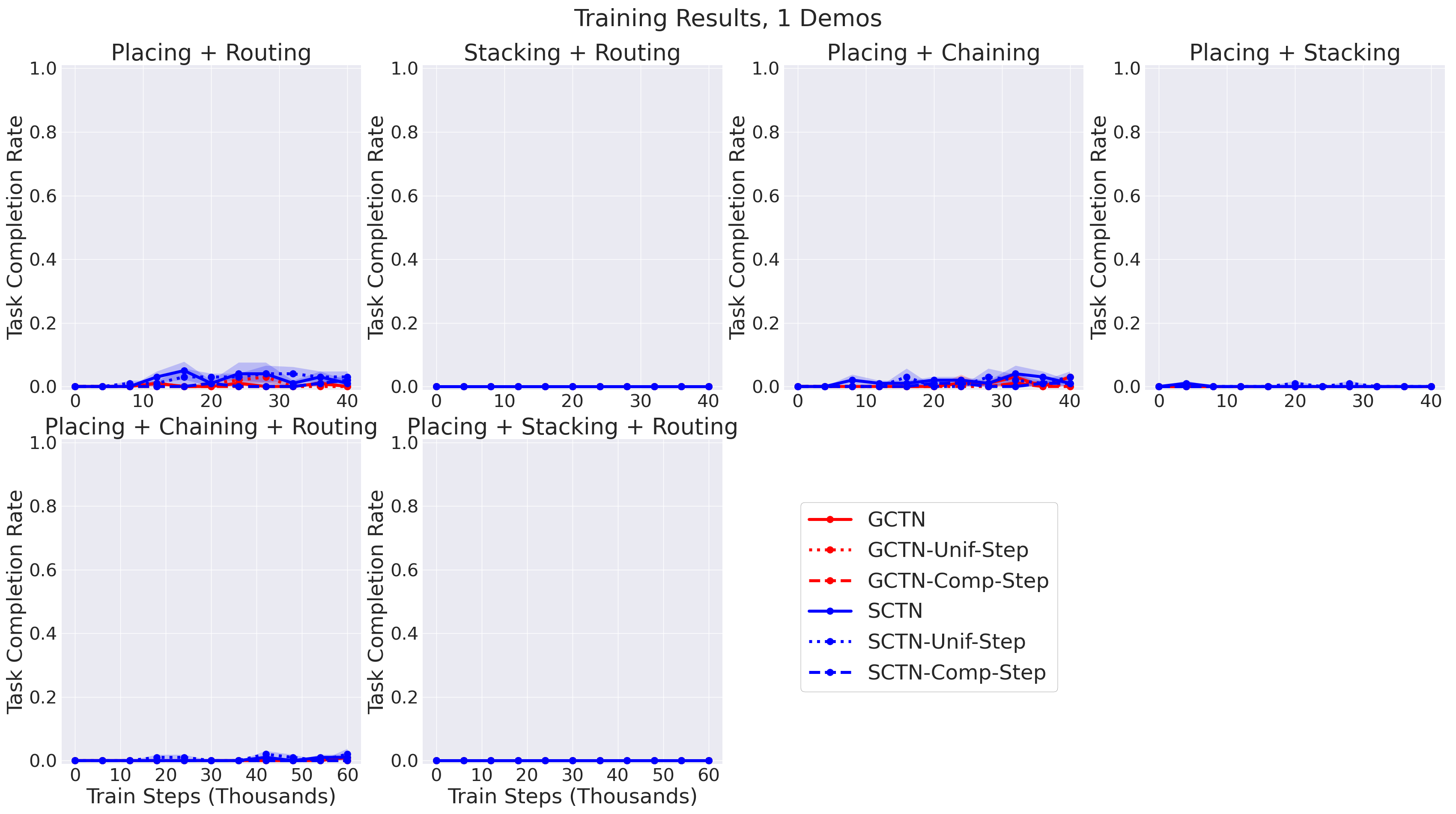}
  \caption{Results for models on the multi-task problems trained on 1 demonstration. The plotting format follows \cref{fig:single-demos}. All double-task problems are trained with 40K training iterations, and triple-task problems with 60K training iterations.}
  \label{fig:multi-demos1}
  \vspace{-1.5em}
\end{figure*}

\begin{figure*}[t]
\centering
  \includegraphics[width=\textwidth]{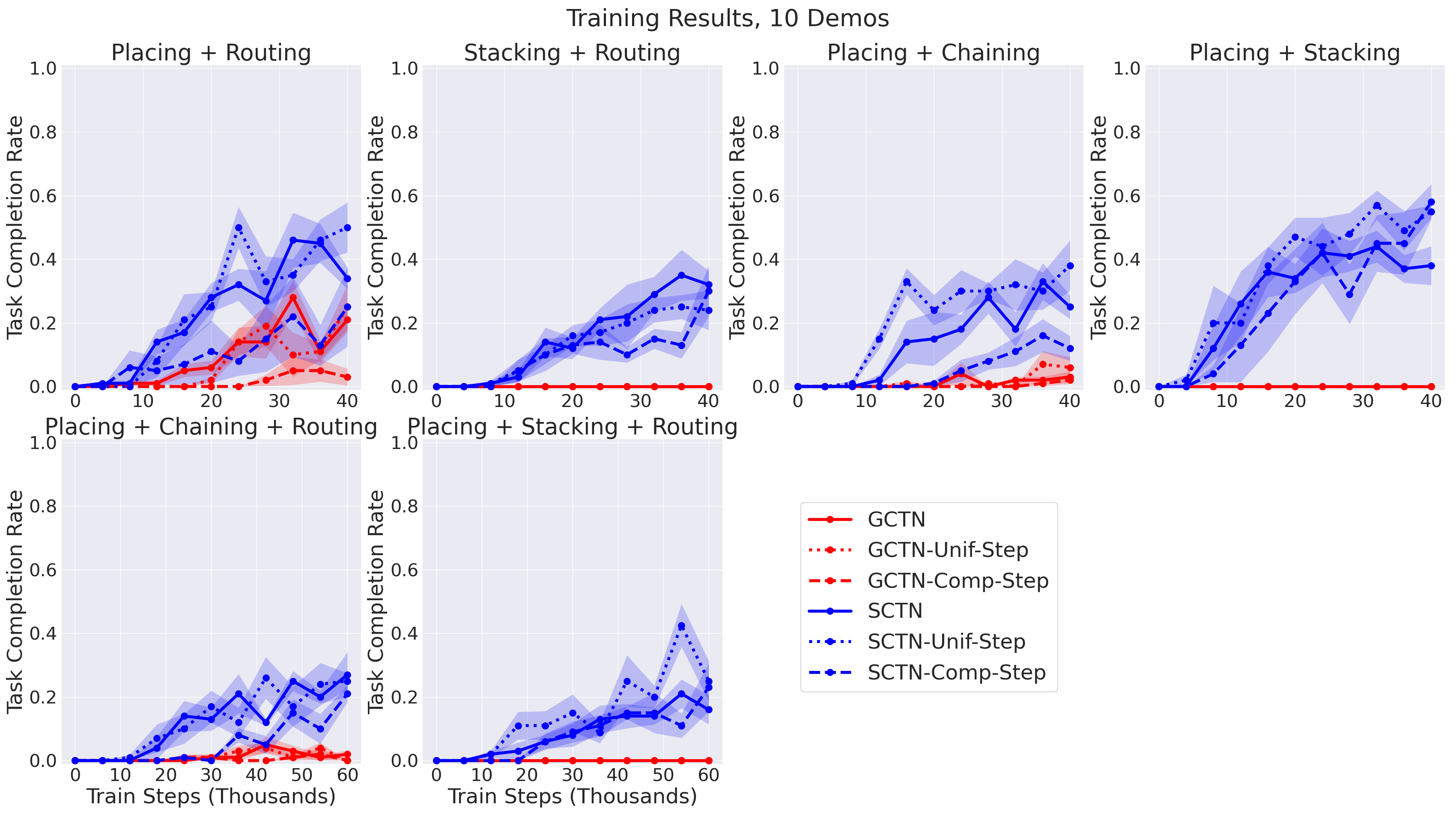}
  \caption{Results for models on the multi-task problems trained on 10 demonstrations. The plotting format follows \cref{fig:multi-demos1}.}
  \vspace{-1.5em}
  \label{fig:multi-demos10}
\end{figure*}

\begin{figure*}[t]
\centering
  \includegraphics[width=\textwidth]{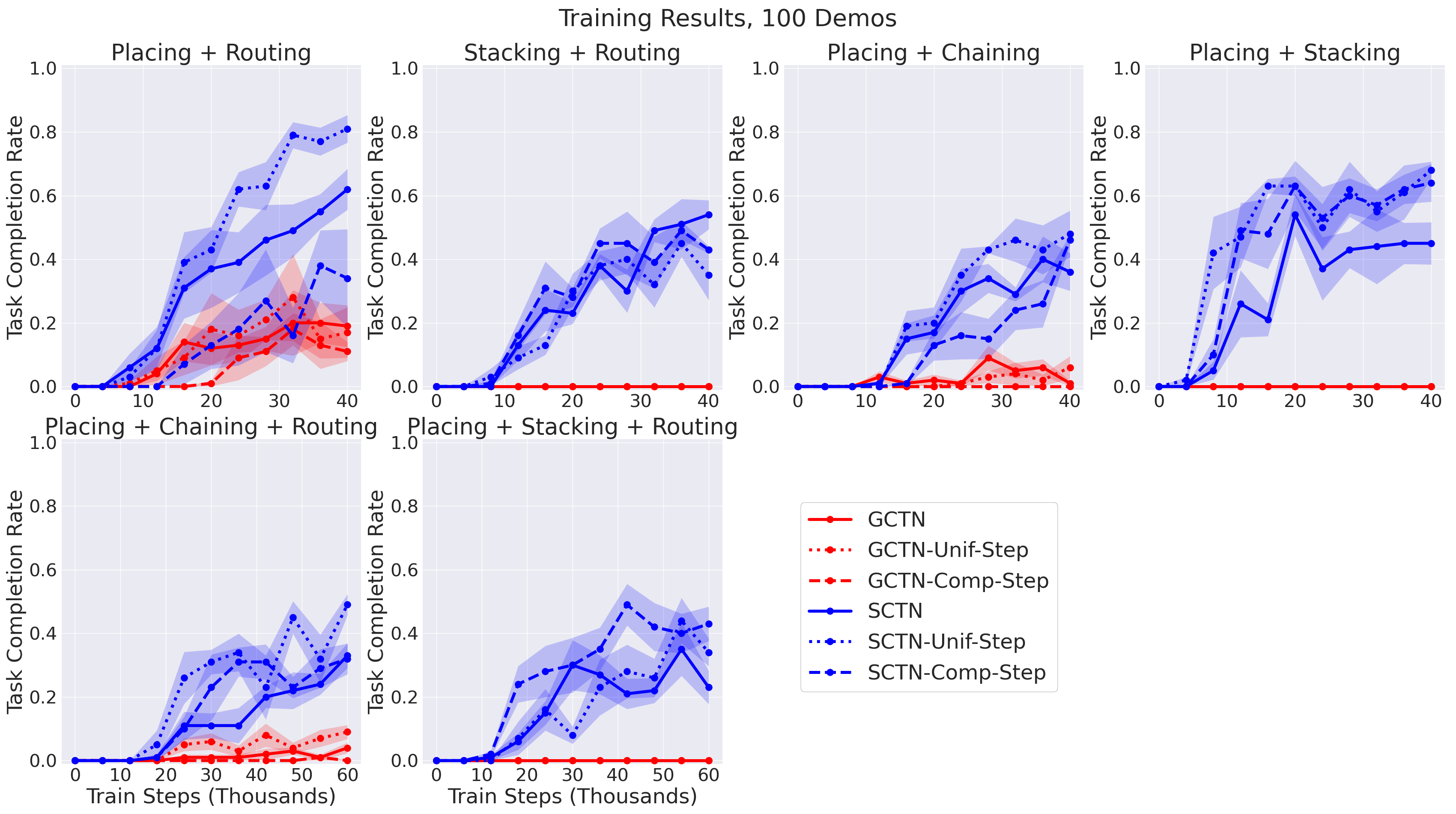}
  \caption{Results for models on the multi-task problems trained on 100 demonstrations. The plotting format follows \cref{fig:multi-demos1}.}
  \vspace{-1.5em}
  \label{fig:multi-demos100}
\end{figure*}

\begin{figure*}[t]
\centering
  \includegraphics[width=\textwidth]{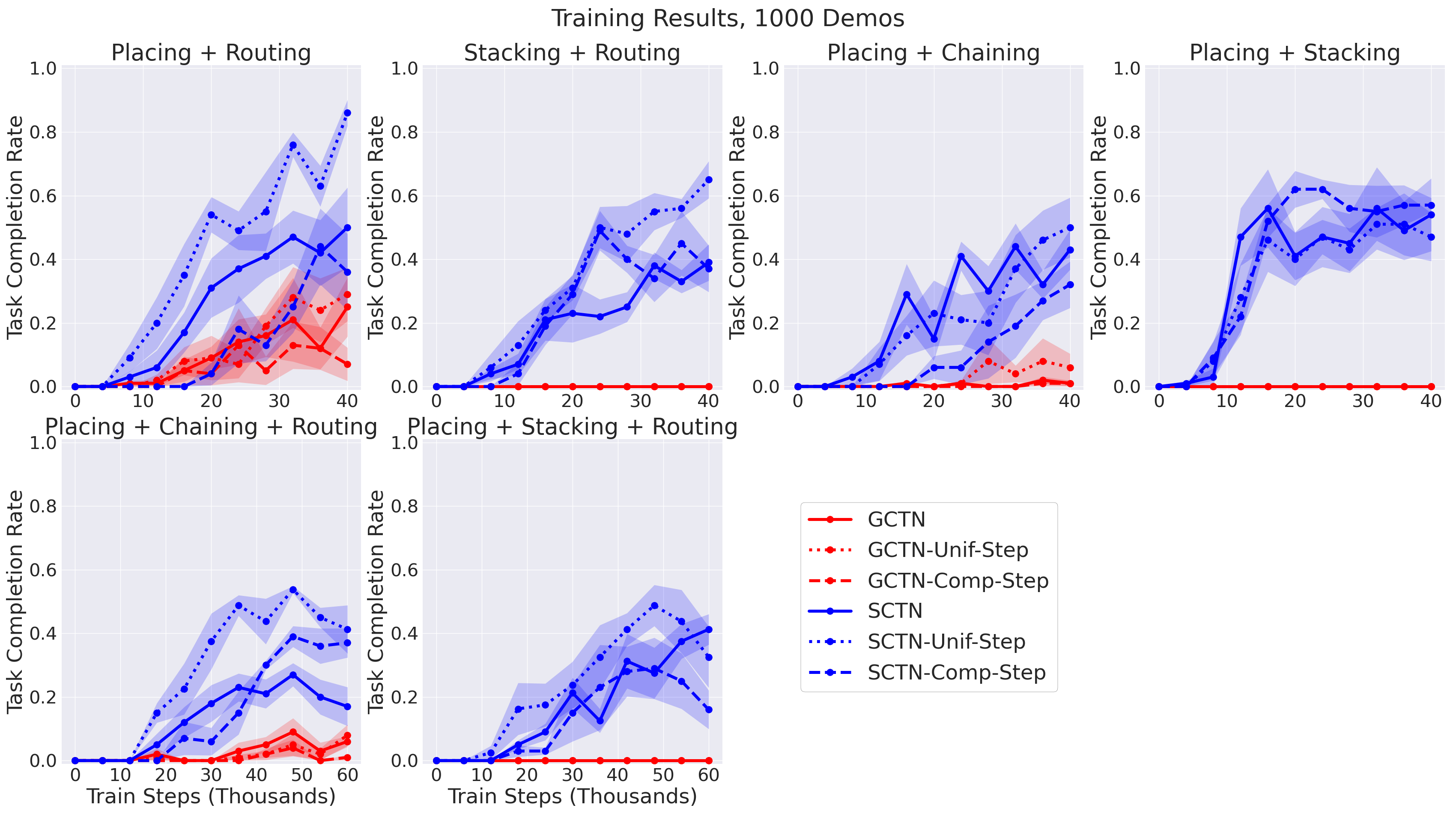}
  \caption{Results for models on the multi-task problems trained on 1000 demonstrations. The plotting format follows \cref{fig:multi-demos1}.}
  \vspace{-1.5em}
  \label{fig:multi-demos1000}
\end{figure*}

The agents are trained with trained for $20\text{K}\times m$ iterations, where $m$ is the number of task modules in a given problem, and one training iteration consists of one gradient update from data of batch size 1.
This means that agents are trained for 20K training iterations for single-task problems, 40K for double-task problems, and 60K for triple-task problems.
At each snapshot occurring every tenth of total training iterations, we evaluate the agents on 20 test episodes.
These 20 episodes are generated using a held-out set of random seeds for testing, which does not overlap with random seeds used for generating demonstrations.
We repeat this training procedure 5 times over 5 different TensorFlow training seeds in order to test consistency of our learning results, totaling $20 \times 5 =100$ metrics for 10 snapshots of the model.
Figs. \ref{fig:multi-demos1}, \ref{fig:multi-demos10}, \ref{fig:multi-demos100} and \ref{fig:multi-demos1000} contain learning curves for multi-task problems. 

\clearpage


\end{appendices}

\end{document}